# A Dynamic Knowledge Distillation Method Based on the Gompertz Curve

YANG Han[1] , QIN Guangjun[1*] ,
[1](Smart City College, Beijing Union University, Beijing 100101, China)
[*](Corresponding author(s). Email(s):zhtguangjun@buu.edu.cn)

**Abstract:**This paper introduces a novel dynamic knowledge distillation framework, Gompertz-CNN, which integrates the Gompertz growth model into the training process to address the limitations of traditional knowledge distillation. Conventional methods often fail to capture the evolving cognitive capacity of student models, leading to suboptimal knowledge transfer. To overcome this, we propose a stage-aware distillation strategy that dynamically adjusts the weight of distillation loss based on the Gompertz curve, reflecting the student's learning progression: slow initial growth, rapid mid-phase improvement, and late-stage saturation.

Our framework incorporates Wasserstein distance to measure feature-level discrepancies and gradient matching to align backward propagation behaviors between teacher and student models. These components are unified under a multi-loss objective, where the Gompertz curve modulates the influence of distillation losses over time. Extensive experiments on CIFAR-10 and CIFAR-100 using various teacher–student architectures (e.g., ResNet50 → MobileNet_v2) demonstrate that Gompertz-CNN consistently outperforms traditional distillation methods, achieving up to 8% and 4% accuracy gains on CIFAR-10 and CIFAR-100, respectively.

## 1. Introduce

In the cutting-edge research field of knowledge distillation, the inherent limitations of the traditional distillation framework[1] have gradually emerged as research deepens. This framework mainly achieves knowledge transfer through soft target matching. However, this method only stays at the superficial imitation of the output layer's probability distribution and fails to deeply explore the complex cognitive reasoning structure contained within the teacher model. This shallow knowledge transfer mechanism leads to a significant decay of higher-order cognitive abilities during the knowledge transfer process, which greatly limits the generalization performance of the student model in complex task scenarios[2].

In response to this key challenge, recent research has actively explored refined distillation methods based on the cognitive process[3]. Among them, the Step-by-Step Distillation paradigm, by explicitly modeling the teacher's chain of thought (CoT), embeds the reasoning trajectory into the knowledge transfer process, achieving hierarchical mapping of cognitive structures. This provides a new theoretical pathway for solving the problem of ability decay in knowledge transfer.

This paper introduces the Gompertz growth model[4] into the knowledge distillation system, constructing a brand-new dynamic knowledge transfer framework.

The Gompertz growth model, with its time-varying parameter characteristics, can accurately characterize the three-stage dynamic features of the student model in the knowledge absorption process: the initial slow growth phase, namely the cognitive foundation phase, during which the student model initially encounters knowledge and accumulates basic cognition; the rapid growth phase, also known as the ability explosion phase, during which the student model's speed of knowledge absorption and ability enhancement significantly increases; and the saturation and stabilization phase, corresponding to the performance convergence phase, during which the student model's performance gradually reaches a stable state. By constructing an adaptive scheduling mechanism based on the Gompertz curve, dynamic weight allocation of multi-stage loss functions is realized, optimizing the temporal process of knowledge distillation. Experimental verification was carried out on the cifar10 and cifar100 datasets, with Resnet50, Resnet34, etc., as teacher models and VGG16, Mobilenet_v2, etc., as student models for preliminary exploratory experiments. The experimental results show that, after comprehensive testing of multiple model combinations, the Gompertz-TCNN framework increased the classification accuracy of the student model by about 8 percentage points and 4 percentage points on the cifar10 and cifar100 datasets, respectively, compared with traditional knowledge distillation methods.

## 2. Research Progress on Knowledge Distillation

The concept of Knowledge Distillation (KD)[5] was first proposed by Geoffrey Hinton et al. in 2015, and its potential in model compression and performance improvement was demonstrated. With the popularity of large models, knowledge distillation is no longer just a tool for "slimming down" models, but a key means of enhancing the performance of small models. By distilling the rich knowledge from large models, the generalization ability of small models can be improved.

**Table 1. Comparison of Knowledge Distillation Methods**

| Distillation Methods | Advantages | Disadvantages |
|---|---|---|
| Knowledge Integration | A multi-purpose student network can reduce deployment costs and improve model utilization. | Compared with single-network prediction for a single target task, the performance of a multi-purpose student network may decline. |
| Multi-teacher Learning | Compared with a single teacher, multiple teachers can usually enable the student network to learn better and richer knowledge. | It is difficult to efficiently integrate the knowledge from each teacher network. |
| Teacher Assistant | Reducing the "generation gap" between teacher and student networks helps the student network to be trained better. | It usually increases the cost of training. |
| Cross-modal Distillation | To achieve small sample learning or semi-supervised learning through cross-modal data sets, reducing the dependence on labeled data | Acquiring synchronized cross-modal data is not an easy task |
| Mutual distillation | Save the time of training teachers online and improve the efficiency of training | It could lead to a situation where "the blind lead the blind" |
| Self-distillation | Save the time of training teachers online and improve the efficiency of training | Lack of rich external knowledge |

As a pivotal technique in machine learning, knowledge distillation is systematically summarized in Table 1-1. Knowledge amalgamation (KA)[6] integrates knowledge from multiple teacher models or tasks into a single student model, enabling it to simultaneously handle diverse tasks.The core of knowledge distillation lies in how student models can leverage knowledge from multiple teacher models to update their parameters, ensuring the final student model can handle tasks originally assigned by these teacher models. Its advantage is that it allows students to learn from broader knowledge sources, though unoptimized implementation may lead to performance degradation. Multi-Teacher Learning in Knowledge Distillation[7] serves as an effective strategy for model compression and knowledge transfer, with its fundamental principle being the utilization of knowledge from multiple teacher models to guide the learning process of the student model.These teacher models can vary in structure, parameters, and performance. Their knowledge is distilled and transferred to student models through a knowledge distillation process. While this approach benefits from diverse knowledge sources, it may face challenges in integration and complexity. The Teacher Assistant (TA) network[8], functioning as an intermediary layer between teacher and student models, serves as a bridge for knowledge transfer. It helps narrow the knowledge gap between teacher and student models, enabling student models to more readily absorb and understand the knowledge from teacher models. This reduces the overhead of teacher networks and facilitates more effective guidance of student networks, even though it may increase time costs for teachers. Cross-modal distillation is a technique that transfers knowledge from one modalities 'teacher model to another modalities' student model. Cross-modal Distillation[9] is a method that leverages complementary insights by transferring data across different models, which can enhance learning efficiency but may be complex and data-intensive. Mutual Distillation[10] refers to the process of knowledge distillation among a group of untrained student models. These student models learn from each other during training, collectively improving performance. As a two-way knowledge-sharing approach, mutual distillation may boost learning rates but faces challenges in maintaining consistent data sharing. Self-Distillation[11-12] involves knowledge distillation within a single model, where the model refines its own knowledge to guide its learning process. Focusing on self-improvement over time, self-distillation enhances efficiency but lacks external influence, making it harder to incorporate new external data.

In the field of knowledge distillation, researchers have conducted multidimensional explorations focusing on model lightweighting and performance preservation. Wu et al.[13] proposed an EKF integrated knowledge distillation framework that generates diverse lightweight models through multi-teacher joint supervision and Dropout-enhanced integration, with its dynamic collaborative reasoning mechanism effectively resolving the contradiction between edge device resource constraints and model accuracy. Liu et al.[14] developed a DFKD method based on dynamic focusing mechanism for industrial inspection scenarios, achieving precise knowledge transfer control through dual weighting factors and temperature-adaptive strategies. This approach effectively retains high-order knowledge from teacher models in lightweight models, significantly improving the accuracy and efficiency of insulator defect detection while optimizing performance and resource consumption on edge devices. Zhu et al.[15] designed a DynamicKD framework from an information theory perspective, which reduces the teacher-student gap by 2.64% on CIFAR100 dataset through real-time adjustment of student model's output entropy distribution. This method achieves significant improvement in knowledge transfer efficiency while maintaining model lightweighting. Singh et al. [16] addressed deployment challenges in NLP by proposing a BERT-inspired knowledge distillation framework, realizing optimal balance between model size and performance on edge devices through multi-teacher collaboration, dynamic temperature adjustment, and multi-stage fine-tuning. The modal weighted knowledge distillation method based on meta-learning proposed by Wang H et al.[17] enables the model to automatically adjust the weight of each mode when data is missing, which can effectively deal with the problem of data loss in multimodal learning and improve the performance of the model in complex situations.

Existing knowledge distillation frameworks overlook the stage-specific characteristics of student models during cognitive development. Specifically, in the early training phase, student models demonstrate low efficiency in absorbing basic semantic features, requiring strong supervisory guidance from teacher models. As training progresses, student models gradually develop foundational representation capabilities. At this stage, supervisory intensity should be progressively reduced to encourage autonomous feature exploration, facilitating a smooth transition from "hand-holding instruction" to "self-directed exploration." This dynamic adjustment mechanism effectively enhances knowledge absorption efficiency, avoids overfitting risks caused by "cramming-style" teaching, improves knowledge transfer efficiency, and strengthens the generalization ability of student models. It provides a novel research perspective for the field of knowledge distillation.

## 3. Gompertz-TCNN

The Gompertz-CNN framework is an innovative architecture that integrates dynamic knowledge distillation with Convolutional Neural Network (CNN), as illustrated in Figure 1. In this framework, the teacher model's is responsible for feature extraction from input data. By combining persistent homology with convolutional operations, this unique combination effectively captures both global structural features and local geometric characteristics of input data, generating teacher feature maps that contain high-order semantic information.

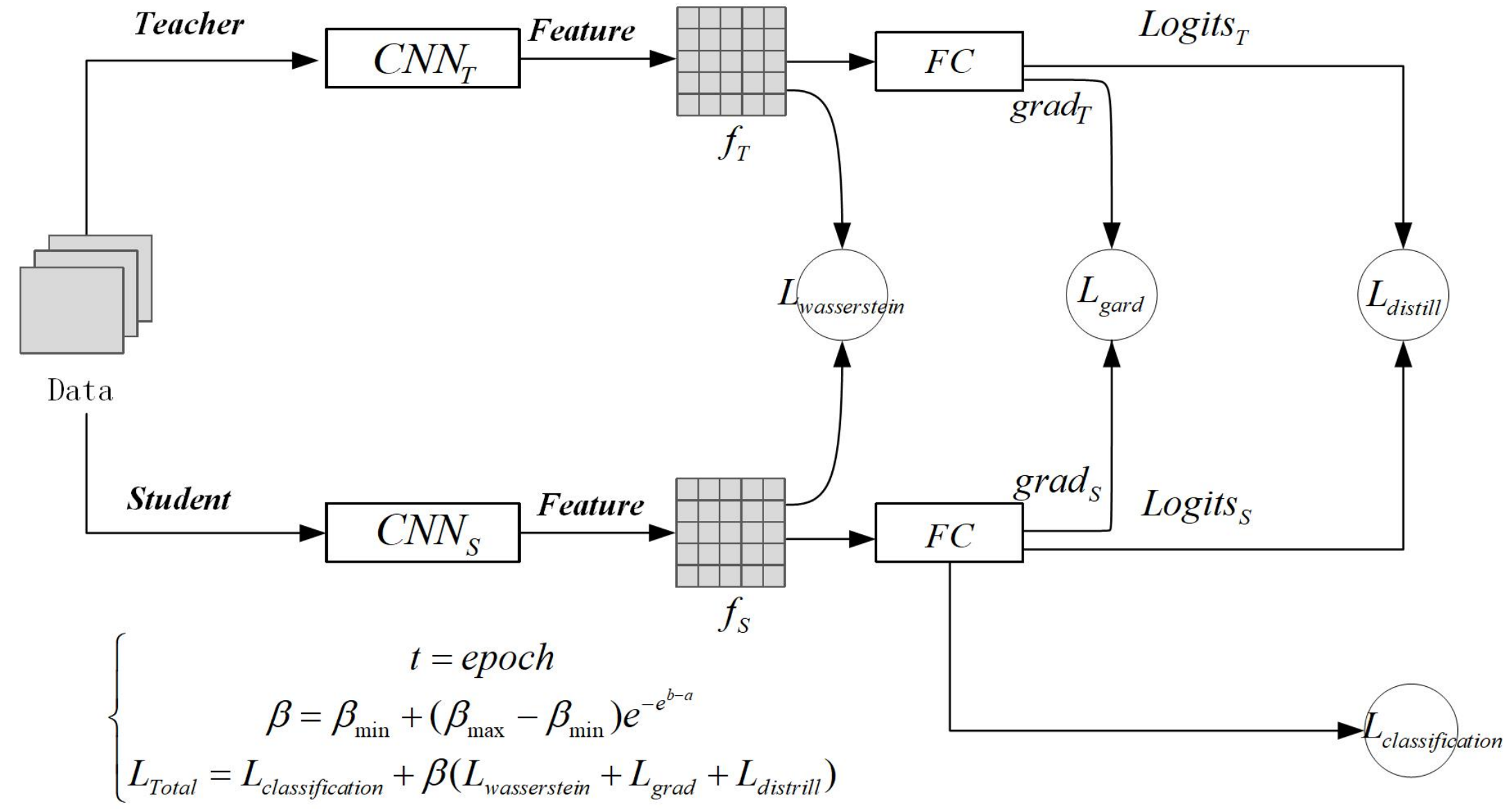

**Figure 1. Gompertz-TCNN model architecture**

The student model employs the CNN architecture for independent learning. During training, it approximates the teacher model's representation capabilities through multi-dimensional loss functions including feature difference loss, gradient matching loss, and distillation loss. Notably, a Gompertz growth curve is introduced to dynamically adjust the distillation loss weight. Throughout the training process, the distillation loss weight exhibits an adaptive variation pattern of "slow growth → rapid enhancement → stable saturation" according to the Gompertz curve. This dynamic adjustment mechanism better accommodates the student model's evolving requirements at different learning stages, thereby optimizing the temporal progression of knowledge transfer.

### 3.1 Feature extraction module

The teacher model employs a specialized CNN architecture to extract features from input images. Convolutional operations focus on extracting local geometric characteristics like edges and textures. After this series of operations, the teacher feature map $f_T \in R^{C\times H\times W}$ is generated before the fully connected layer performs classification tasks. This map contains both high-level semantic representations of image data.

The feature extraction process of student models is similarly based on the CNN architecture, which can adopt either homogeneous or heterogeneous configurations with teacher models. The student model generates feature maps$f_S \in R^{C\times H\times W}$ , serving as approximate representations or knowledge extraction results of the teacher's feature maps. These maps reflect the student model's understanding hierarchy of image data. This design enables student models to progressively enhance their data comprehension and processing capabilities while emulating the teacher model.

### 3.2 Wasserstein distance measure and feature loss

To precisely quantify the differences between teacher and student feature maps, we employ the Wasserstein distance (also known as Earth-Mover's Distance, EMD) [18] as a metric. This metric conducts an in-depth comparison of representation between teacher models and student models within the feature space. By calculating the minimum "cost" required to transform one feature map into another, it measures the degree of difference between the two feature maps.

Specifically, in the computational process, the feature maps are treated as probability distributions. The Wasserstein distance effectively captures the differences between two distributions by considering not only variations in feature values but also their spatial distribution patterns. By calculating the Wasserstein distance between two feature maps, we construct the corresponding loss term $L_{Wasserstein}$ as shown in Formula 1. This loss term precisely characterizes the degree of divergence between the student model and the teacher model in feature extraction dimensions, becoming a core optimization objective that requires prioritized minimization during knowledge distillation. Through continuous optimization of this loss term, the student model can better approximate the teacher model in feature extraction.

$$L_{\mathrm{wasser}stein} = \inf_{r\in\Pi(f_T,f_S)} E_{x,y\sim\gamma}[\|x-y\|] \tag{1}$$

### 3.3 Gradient matching optimization strategy

During the training of deep learning models, accurately extracting gradient information from feature maps of two different models is of paramount importance. This critical operation not only directly impacts the accuracy of model parameter updates but also serves as a crucial foundation for optimizing model performance and enhancing feature learning effectiveness. Given that the architectures of teacher models and student models may differ, their gradient channels might also vary. To achieve effective channel-level alignment between teacher and student model gradients, this study designs a 1×1 convolutional layer to perform channel remapping on the teacher model's gradient, ensuring its channel count matches that of the student model's gradient.

For the teacher model gradient $grad_T$ and student model gradient $grad_S$ after channelized processing, Euclidean distance and cosine similarity are respectively used for measurement. Euclidean distance primarily measures the absolute difference between gradient vectors, intuitively reflecting the numerical gap between them. Cosine similarity focuses on assessing directional consistency by calculating the cosine of the angle between two vectors to determine their alignment. As shown in Formula 2 and 3:

Euclidean distance measures the absolute difference between gradient vectors:

$$D_{euclidean} = \| grad_T - grad_S \|_2 \tag{2}$$

Cosine similarity measures gradient direction consistency:

$$S_{\cos ine} = \frac{grad_T \bullet grad_S}{\| grad_T \|_2 \| grad_S \|_2} \tag{3}$$

To balance the impact of gradient magnitude differences and direction consistency in model training, a weight coefficient r is introduced for adjustment, as shown in Equation 4. By properly setting this coefficient, the student model can comprehensively learn from the teacher model's characteristics during optimization. This approach not only replicates the teacher model's forward propagation results but also deeply incorporates its backpropagation features, thereby significantly enhancing the performance of the student model.

$$L_{grad} = r \cdot D_{euclidean} + (1 - r) \cdot (1 - S_{\text{cosine}}) \tag{4}$$

### 3.4 Dynamic weight scheduling based on Gompertz curve

The Gompertz curve can accurately depict the dynamic evolution process of "slow initial growth, accelerated mid-term growth, and later slowing growth", whose standard form is shown in Formula 5. Here, *K* represents the asymptotic upper limit of the curve, which indicates the ultimate state that can be reached after a long period of evolution.

$$y = k . e^{-e^{b-ct}} \tag{5}$$

This study conducts adaptive optimization of the Gompertz curve based on practical requirements for knowledge distillation, establishing a dynamic adjustment mechanism for distillation loss parameters. Specifically, $\beta_{max}$ and $\beta_{min}$ are defined as the maximum and minimum boundaries of distillation losses, respectively, with the constraints $\beta_{min} < \beta_{max}$、$\beta_{max} = 1.0$ and $\beta_{min} = 0.1$. The training iteration count (epoch) is treated as the time variable t, while b serves as the growth rate parameter that controls both the curve's progression speed and shape. The transformed formula is presented in Equations 6:

$$\beta = \beta_{min} + (\beta_{max} - \beta_{min}) e^{-e^{b-ct}} \tag{6}$$

This formula causes the distillation loss weight βto gradually increase during the initial training phase. This is because at the beginning of training, the student model's understanding and knowledge absorption capacity are limited. Excessive distillation loss weight may lead to over-reliance on the teacher model, thereby suppressing its own learning ability. As training progresses into the mid-phase, the weight β rises rapidly. At this stage, the student model has accumulated sufficient foundational knowledge to absorb the teacher model's information more effectively. Increasing the distillation loss weight facilitates accelerated knowledge transfer. In the late training phase, the weight β stabilizes. At this point, the student model's performance gradually converges. A stable distillation loss weight ensures fine-tuning based on existing knowledge while avoiding performance degradation from excessive adjustments. This dynamic adjustment mechanism effectively aligns with the cognitive development patterns of the student model.

### 3.5 Total loss function

The Gompertz-CNN framework is an innovative architecture that integrates dynamic knowledge distillation with CNN, as illustrated in Figure 1. In this framework, the teacher model's is responsible for feature extraction from input data. By combining persistent homology with convolutional operations, this unique combination effectively captures both global structural features and local geometric characteristics of input data, generating teacher feature maps that contain high-order semantic information.

Distillation loss serves as the cornerstone for knowledge transfer, where τ denotes the distillation temperature parameter, and the softmax activation function converts model outputs into probability distributions. Through this mechanism, student models can learn from the soft-target guidance provided by the teacher model, thereby acquiring the rich knowledge embedded within the teacher's architecture.

Meanwhile, by integrating the feature loss measured through Wasserstein distance with the gradient matching loss, we enhance both feature consistency and gradient alignment. These loss terms impose constraints on student models from multiple perspectives, ensuring they better approximate the teacher model in aspects such as feature extraction and gradient learning.

The final total loss function for the Gompertz-CNN method is defined in Equation 7. Here, β acts as a distilled loss weight coefficient that plays a crucial balancing role during training, effectively adjusting the optimization weights between classification losses and multi-component distilled losses. Through this joint optimization strategy, the model achieves comprehensive performance enhancement, enabling the student model to demonstrate superior results across multiple dimensions.

$$L_{total} = L_{classification} + \beta \cdot (L_{wasserstein} + L_{grad} + L_{distill}) \quad (7)$$

(1) Classification loss $L_{classification} = CrossEntropy(y, Log_S)$ reflects the performance of the model in the classification task;

(2) Distillation loss $L_{distill} = KDDiv(\sigma(Log_T/\tau), \sigma(Log_S/\tau))$ to achieve knowledge transfer; where $\tau$ is the distillation temperature parameter; $\sigma$ represents the softmax activation function.

## 4. Algorithm design

In the training optimization of deep neural networks, the Gompertz-CNN algorithm enhances student model performance through a dynamic knowledge distillation mechanism to adapt to complex task requirements. The following is a detailed description of the algorithm.

**Algorithm 1: Gompertz-CNN training algorithm**

**Input: Dataset** $D = \{(xi, yi)\}_{i=1}^{N}$ **, Growth rate** $b$**, Training rounds** $t$

**Output: The accuracy of the student model for image classification after training**

1. **Initialization: Initialize the parameters of teacher model and student model;**
2. **Training cycle (epoch cycle):**

*for* $i = 1$ to $t$

*for(xi, yi) in D do*

*(1) feature extraction:*

- *The teacher model is used to extract features of xi, and the feature map* $f_T \in R^{C \times H \times W}$ *of the teacher model is obtained;*
- *The student model is used to extract features of xi, and the feature map* $f_S \in R^{C' \times H' \times W'}$ *of the student model is obtained;*

*(2) Calculate loss value*

- *Loss of feature differences: Calculate the Wasserstein distance between the teacher's feature map* $f_T \in R^{C \times H \times W}$ *and the student's feature map* $f_S \in R^{C' \times H' \times W'}$ *to obtain the feature difference loss*

$$L_{wasserstein} = inf_{r \in \prod(f_t, f_s)}(E_{x,y \sim \psi}[\|x - \psi\|]) ;$$

- *Gradient matching loss:*
  - *Obtain the gradient* $grad_T$ *and* $grad_S$ *of teacher model and student model before classification layer.*
  - *The 1×1 convolution layer is used to perform channel remapping on* $grad_T$ *to make its channel number consistent with* $grad_S$ *, and grade* $grad_T^{re}$ *is*

*obtained;*

- *Calculate the Euclidean distance* $D_{euclidean}$ *and cosine similarity* $grad_T^{re}$ *between* $grad_T^{re}$ *and* $grad_S$*;*
- *Loss of computing ladder distribution, r is the weight coefficient*

$$L_{grad} = r \cdot D_{euclidean} + (1-r) \cdot (1-S_{cosine})$$

- *distillation loss*
  - *Calculate the logit of the teacher model and the student model* $logit(T)$, $logit(S)$
  - *The probability distribution is obtained by processing the logit with the softmax function combined with the distillation temperature T*

$$P_T = soft\max(Logit(T)/\tau) \text{ 和 } P_S = softmax(Logit(S)/\tau)$$

  - *Calculate distillation losses* $L_{distill} = KLDiv(P_T, P_S)$
- *Classification loss: According to the prediction result* $S(xi)$ *of the student model and the real label* $yi$*, the classification loss is calculated*

*(3) Dynamic weight calculation: Calculate the distillation loss weight* $\beta$ *for the current training round based on the Gompertz curve formula.*

*(4) Calculate total losses*

$$L_{total} = L_{classification} + \beta \cdot (L_{wasserstein} + L_{grad} + L_{distill})$$

*(5) Model update: The parameters of the student model are updated using the optimizer according to the total loss.*

*End for*

*End for*

3. **Return results: Return the trained student model**

By dynamically adjusting the weight of distillation loss and integrating multi-dimensional loss function, this algorithm can effectively improve the performance of student model and provide support for the application of deep model in resource-constrained scenarios.

## 5. Experimental analysis

### 5.1 Experimental design

In model selection, we conducted comprehensive evaluations and screening to identify a series of high-performance CNN architectures with diverse structures, ensuring the experimental results demonstrate comprehensiveness and representativeness. The selected models include ResNet50, VGG16, ResNet34, and MobileNet_v2. These models not only excel in tasks such as image classification and object detection but also feature unique network architectures and optimization strategies. This enables in-depth analysis of knowledge transfer efficiency and performance differences between models during the knowledge distillation process.

During the experimental process, the proposed Gompertz-CNN method was applied to conduct detailed training and evaluation of different models. In the data preprocessing phase, the experimental dataset underwent rigorous standardization including normalization operations, which mapped image pixel values into a specific range to accelerate model convergence. During model training, various hyperparameters were carefully adjusted, such as learning rates and optimizer selection. Different initial values for learning rates were set with a decay strategy implemented: a higher learning rate was applied in the early training phase to speed up convergence, while the rate was gradually reduced as training progressed to prevent model oscillation in later stages.

### 5.2 experimental result

Table 2 shows the classification performance comparison between traditional knowledge distillation method and Gompertz-CNN dynamic distillation method on CIFAR-10 and CIFAR-100 benchmark data sets when homogeneous and heterogeneous teacher model and student model architectures are adopted respectively.

**Table 2 Comparison of classification accuracy**

| Dataset | Teacher Model | Student model | Teacher model | Student model | Traditional knowledge distillation | Gompertz -CNN |
|---|---|---|---|---|---|---|
| cifar10 | Resnet50 | VGG16 | 0.8466 | 0.9121 | 0.6991 | 0.9399 |
| | Resnet50 | Resnet34 | 0.8466 | 0.8673 | 0.9181 | 0.9231 |
| | Resnet50 | Mobilenet_v2 | 0.8466 | 0.7611 | 0.9145 | 0.9574 |
| | Resnet34 | Mobilenet_v2 | 0.8673 | 0.7611 | 0.9152 | 0.9563 |
| cifar100 | Resnet50 | VGG16 | 0.6234 | 0.8600 | 0.7138 | 0.7399 |
| | Resnet50 | Resnet34 | 0.6234 | 0.5583 | 0.6940 | 0.6722 |
| | Resnet50 | Mobilenet_v2 | 0.6234 | 0.2437 | 0.7043 | 0.7709 |
| | Resnet34 | Mobilenet_v2 | 0.5583 | 0.2437 | 0.7089 | 0.7741 |

This study utilizes two widely adopted machine learning datasets, CIFAR-10 and CIFAR-100, to evaluate the performance of different teacher-student model combinations under traditional knowledge distillation methods and the Gompertz-CNN approach. When employing ResNet50 or ResNet34 as teacher models with various student model configurations across datasets, the Gompertz-CNN method consistently outperforms traditional approaches in classification accuracy.

Comprehensive evaluations reveal that on CIFAR-10, the Gompertz-CNN method achieves an average improvement of approximately 8 percentage points compared to conventional knowledge distillation methods, while on CIFAR-100, this enhancement reaches about 4 percentage points. These results demonstrate that the Gompertz-CNN method effectively boosts classification accuracy across multiple datasets and model combinations, outperforming traditional knowledge distillation methods with significant advantages.

## 6. Methodological analysis

The radar chart can comprehensively display the performance of the model under various evaluation indicators. By comparing the graphic area and shape of different methods in the radar chart, the comprehensive advantages of Gompertz-CNN method can be visually seen.

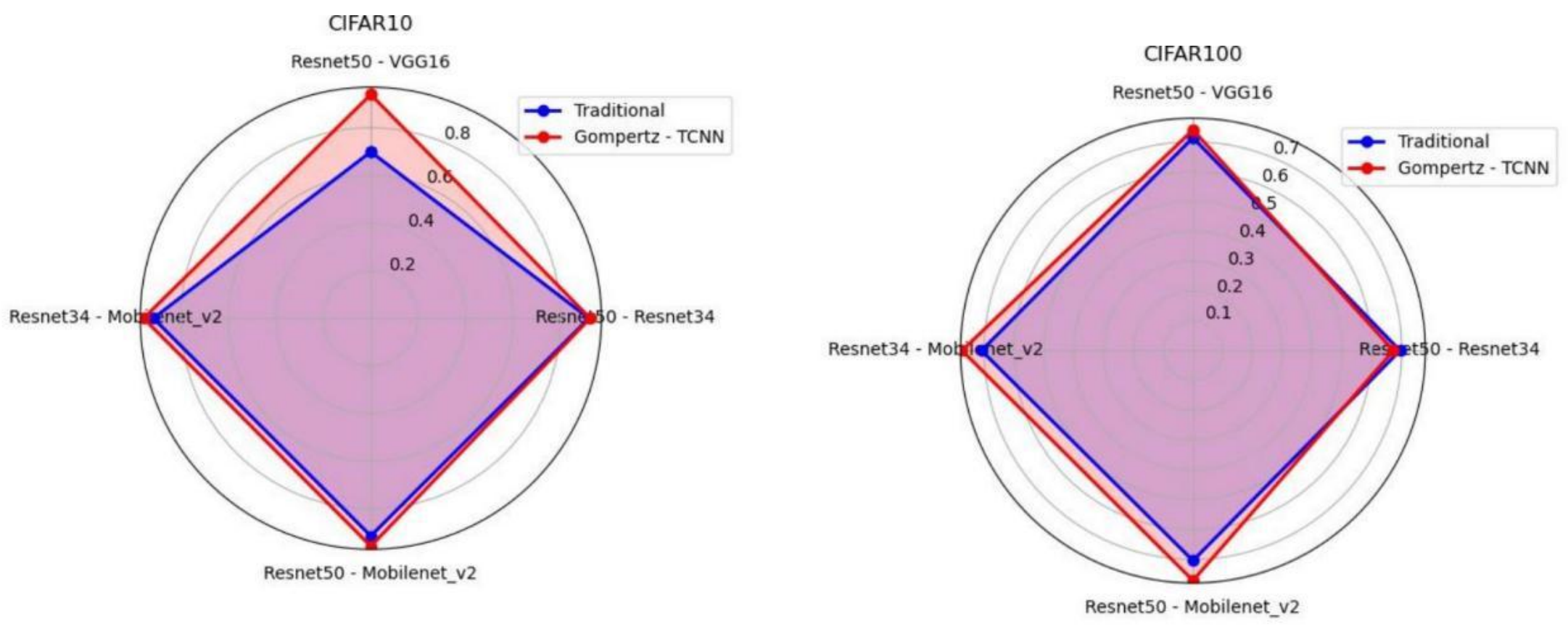

**(a) Performance comparison on CIFAR10 data set** **(b) Performance comparison on CIFAR100 data set**

**Figure 2 Comparison of traditional methods and Gompertz-TCNN methods visualized on radar chart**

The analysis of these two radar plots reveals that the Gompertz-CNN method (red area) demonstrates a larger classification accuracy area than traditional knowledge distillation methods (blue area) on the CIFAR10 dataset. This indicates superior performance of Gompertz-CNN in teacher-student model combinations including Resnet50-VGG16, ResNet50-ResNet34, ResNet50-MobileNet_v2, and ResNet34-MobileNet_v2. Notably, the Gompertz-TCNN method shows significant accuracy improvements over conventional approaches in specific configurations like ResNet50-Mobilenet_v2, highlighting its effective enhancement of student model performance under particular network architectures.

The Gompertz-CNN method maintains its dominance on the CIFAR100 dataset: While performance variations exist across different model combinations, it consistently outperforms traditional knowledge distillation methods (blue regions) in most configurations. For instance, when paired with ResNet50-MobileNet_v2 or ResNet34-MobileNet_v2, it demonstrates a larger enclosed area (indicating higher classification accuracy). Although slightly inferior to conventional approaches in ResNet50-ResNet34 combinations, the Gompertz-CNN method still shows significant advantages in enhancing student model accuracy across the CIFAR100 dataset.

Based on the results of the two data sets, compared with the traditional knowledge distillation method, Gompertz-CNN method has more advantages in improving the classification accuracy of student models, especially on CIFAR10 dataset.

## 7. Summary

The paper proposes a Gompertz curve-based dynamic knowledge distillation method to resolve the contradiction between efficient model construction and resource consumption. The research focuses on improving knowledge transfer efficiency in deep learning: First, process supervision is introduced to measure the differences between teacher and student model feature maps using Wasserstein distance, with gradient evaluation assessing decision differences in fully connected layers. Second, the Gompertz curve dynamically adjusts weight parameters during knowledge distillation to enhance student model performance. Experimental validations on CIFAR-10 and CIFAR-100 datasets demonstrate this approach, with preliminary exploratory experiments using ResNet50 and ResNet34 as teacher models against VGG16 and MobileNet_v2 as student models. Results show that after comprehensive testing of multiple model combinations, the Gompertz-CNN framework improves classification accuracy by approximately 8 percentage points and 4 percentage points for student models on CIFAR-10 and CIFAR-100 datasets, respectively, compared to traditional knowledge distillation methods.

We will continue to conduct in-depth research on optimization strategies and parameter adjustment mechanisms for loss functions. By systematically analyzing the synergistic effects and complementary characteristics among different loss functions, we aim to explore more efficient weight allocation schemes. Simultaneously, we will focus on optimizing the dynamic knowledge distillation framework based on Gompertz curves, delving into the relationship between their nonlinear characteristics and knowledge transfer efficiency. Through improvements in adaptive curve parameter adjustment algorithms, we enhance the framework's flexibility. Building on this foundation, we will further improve deep learning models' adaptability to complex multi-dimensional data, including challenges such as handling imbalanced data distributions and significant differences in feature dimensions. By integrating multi-task joint training with transfer learning, we significantly enhance model performance accuracy and robustness across various real-world application scenarios.

## Rreference

[1] Georgiou T, Liu Y, Chen W, et al. A survey of traditional and deep learning-based feature descriptors for high dimensional data in computer vision. Int J Multimed Info Retr, 2020, 9(3): 135～170.

[2] Sun S, Cheng Y, Gan Z, et al. Patient Knowledge Distillation for BERT Model

Compression.Proceedings of the 2019 Conference on Empirical Methods in Natural Language Processing and the 9th International Joint Conference on Natural Language Processing(EMNLP-IJCNLP). Hong Kong, China: Association for Computational Linguistics, 2019. 4322～4331.

[3] Hsieh C Y, Li C L, Yeh C kuan, et al. Distilling Step-by-Step! Outperforming Larger Language Models with Less Training Data and Smaller Model Sizes. Findings of the Association for Computational Linguistics: ACL 2023. Toronto, Canada: Association for Computational Linguistics, 2023. 8003～8017.

[4] On the nature of the function expressive of the law of human mortality, and on a new mode of determining the value of life contingencies. In a letter to Francis Baily, Esq. F. R. S. &c. By Benjamin Gompertz, Esq. F. R. S. Proc R Soc Lond, 1833, 2: 252～253.

[5] Szegedy C, Liu W, Jia Y, et al. Going Deeper with Convolutions. arXiv, 2014.

[6] Hinton G, Vinyals O, Dean J. Distilling the Knowledge in a Neural Network. arXiv, 2015.

[7] Shen C, Wang X, Song J, et al. Amalgamating Knowledge towards Comprehensive Classification. arXiv, 2018.

[8] Liu Y, Zhang W, Wang J. Adaptive Multi-Teacher Multi-level Knowledge Distillation. Neurocomputing, 2020, 415: 106～113.

[9] Mirzadeh S I, Farajtabar M, Li A, et al. Improved Knowledge Distillation via Teacher Assistant. arXiv, 2019.

[10] Gupta S, Hoffman J, Malik J. Cross Modal Distillation for Supervision Transfer. arXiv, 2015.

[11] Chen W, Li S, Huang C, et al. Mutual Distillation Learning Network for Trajectory-User Linking. arXiv, 2022.

[12] Zhang L, Song J, Gao A, et al. Be Your Own Teacher: Improve the Performance of Convolutional Neural Networks via Self Distillation. arXiv, 2019.

[13] Wu S, Li Y, Xu Y, et al. EKDF: An Ensemble Knowledge Distillation Framework for Robust Collaborative Inference on Heterogeneous Edge Devices.2023 19th International Conference on Mobility, Sensing and Networking (MSN). Nanjing, China: IEEE, 2023. 191～198.

[14] ]Liu B, Jiang W. DFKD: Dynamic Focused Knowledge Distillation Approach for Insulator Defect Detection. IEEE Trans Instrum Meas, 2024, 73: 1～16.

[15] Zhu S, Shang R, Yuan B, et al. DynamicKD: An effective knowledge distillation via dynamic entropy correction-based distillation for gap optimizing. Pattern Recognition, 2024,153: 110545.

[16] Singh S, Sharma K, Karna B K, et al. A New BERT-Inspired Knowledge Distillation Approach Toward Compressed AI Models for Edge Devices.Rao U P, Alazab M, Gohil B N, et al. Security, Privacy and Data Analytics:1049. Singapore: Springer Nature Singapore, 2023. 105～117.

[17] Wang H, Hassan S, Liu Y, et al. Meta-Learned Modality-Weighted Knowledge Distillation

for Robust Multi-Modal Learning with Missing Data. arXiv, 2025.

[18] Arjovsky M, Chintala S, Bottou L. Wasserstein GAN. arXiv, 2017